\documentclass[11pt,english]{article}
\usepackage{amsmath}
\usepackage{graphicx}
\usepackage{amssymb}
\usepackage{esint}
\usepackage[authoryear]{natbib}
\usepackage{amsfonts,natbib}
\usepackage{url}
\usepackage{tabularx}
\usepackage{dsfont}
\usepackage{epsf}
\usepackage{algorithmic,algorithm}

\newcommand{\BEAS}{\begin{eqnarray*}}
\newcommand{\EEAS}{\end{eqnarray*}}
\newcommand{\BEA}{\begin{eqnarray}}
\newcommand{\EEA}{\end{eqnarray}}
\newcommand{\BEQ}{\begin{equation}}
\newcommand{\EEQ}{\end{equation}}
\newcommand{\BIT}{\begin{itemize}}
\newcommand{\EIT}{\end{itemize}}
\newcommand{\BNUM}{\begin{enumerate}}
\newcommand{\ENUM}{\end{enumerate}}

\newcommand{\BA}{\begin{array}}
\newcommand{\EA}{\end{array}}
\newcommand{\BC}{\begin{center}}
\newcommand{\EC}{\end{center}}

\newcommand{\eg}{{\it e.g.}}

\newcommand{\argmin}{\mathop{\rm argmin}}

\newcounter{exno}

{\begin{quote}}{\end{quote}}

\makeatother

\usepackage{dsfont}

\def\ba{\mathbf{a}}
\def\bm{\mathbf{m}}

\def\bd{\mathbf{d}}

\def\bA{\mathbf{A}}

\def\bb{\mathbf{b}}
\def\bx{\mathbf{x}}

\def\bg{\mathbf{g}}

\usepackage{enumitem}
\usepackage{enumerate}
\usepackage{amsmath}

\def\RR{\mathbb{R}}

\def\Ecal{\mathcal{E}}

\def\Mcal{M_+^b(\Xcal)}

\def\Hcal{\mathcal{H}}

\def\Ical{\mathcal{I}}

\def\one{\mathds{1}}

\def\OMIT#1{}

\DeclareMathOperator{\defi}{def}
\DeclareMathOperator{\tr}{tr}
 
\DeclareMathOperator{\defeq}{\overset{\defi}{=}}

\usepackage{enumerate,multirow}

\providecommand{\abs}[1]{\lvert#1\rvert} 
\providecommand{\norm}[1]{\lVert#1\rVert}

\usepackage{tabularx}

\makeatletter
\newif\if@borderstar
\def\bordermatrix{\@ifnextchar*{%
  \@borderstartrue\@bordermatrix@i}{\@borderstarfalse\@bordermatrix@i*}%
}
\def\@bordermatrix@i*{\@ifnextchar[{%
  \@bordermatrix@ii}{\@bordermatrix@ii[()]}
}
\def\@bordermatrix@ii[#1]#2{%
  \begingroup
    \m@th\@tempdima8.75\p@\setbox\z@\vbox{%
      \def\cr{\crcr\noalign{\kern 2\p@\global\let\cr\endline }}%
      \ialign {$##$\hfil\kern 2\p@\kern\@tempdima & \thinspace %
      \hfil $##$\hfil && \quad\hfil $##$\hfil\crcr\omit\strut %
      \hfil\crcr\noalign{\kern -\baselineskip}#2\crcr\omit %
      \strut\cr}}%
    \setbox\tw@\vbox{\unvcopy\z@\global\setbox\@ne\lastbox}%
    \setbox\tw@\hbox{\unhbox\@ne\unskip\global\setbox\@ne\lastbox}%
    \setbox\tw@\hbox{%
      $\kern\wd\@ne\kern -\@tempdima\left\@firstoftwo#1%
        \if@borderstar\kern2pt\else\kern -\wd\@ne\fi%
      \global\setbox\@ne\vbox{\box\@ne\if@borderstar\else\kern 2\p@\fi}%
      \vcenter{\if@borderstar\else\kern -\ht\@ne\fi%
        \unvbox\z@\kern-\if@borderstar2\fi\baselineskip}%
        \if@borderstar\kern-2\@tempdima\kern2\p@\else\,\fi\right\@secondoftwo#1 $%
    }\null \;\vbox{\kern\ht\@ne\box\tw@}%
  \endgroup
}
\makeatother

\usepackage{graphicx} 
\usepackage{subfigure}

\def\ve{\varepsilon}

\def\bx{\textbf{x}}

\def\ba{\mathbf{a}}

\def\bA{\mathbf{A}}
\def\bB{\mathbf{B}}

\def\bb{\mathbf{b}}

\def\RR{\mathbb{R}}

\def\Mcal{\mathcal{D}}

\def\Mcal{\mathcal{M}}

\def\Hcal{\mathcal{H}}

\def\Ical{\mathcal{I}}

\def\one{\mathds{1}}

\def\OMIT#1{}

\newcommand{\dotprod}[2]{\ensuremath{\langle #1 , #2\,\rangle}}

\usepackage{hyperref}

\DeclareMathOperator{\Ex}{Ex}
\def\ExP{\Ex(r,c)}
\def\cs{\Sigma_{d-1}} 

\mathchardef\minus="002D

\bibpunct{(}{)}{,}{a}{}{;}
\makeatother

\usepackage{babel}

\begin{document}

\title{Ground Metric Learning}

\author{
Marco Cuturi, David Avis \\
Kyoto University\\
\texttt{\{mcuturi,davis\}@i.kyoto-u.ac.jp}
}

\maketitle
\begin{abstract}
Transportation distances have been used for more than a decade now in machine learning to compare histograms of features. They have one parameter: the \emph{ground metric}, which can be \emph{any} metric between the features themselves. As is the case for all parameterized distances, transportation distances can only prove useful in practice when this parameter is carefully chosen. To date, the only option available to practitioners to set the ground metric parameter was to rely on \emph{a priori} knowledge of the features, which limited considerably the scope of application of transportation distances. We propose to lift this limitation and consider instead algorithms that can \emph{learn} the ground metric using only a training set of labeled histograms. We call this approach ground metric learning. We formulate the problem of learning the ground metric as the minimization of the difference of two polyhedral convex functions over a convex set of distance matrices. We follow the presentation of our algorithms with promising experimental results on binary classification tasks using GIST descriptors of images taken in the Caltech-256 set.
\end{abstract}

\section{Introduction}
We consider in this paper the problem of supervised metric learning on normalized histograms. Normalized histograms arise frequently in natural language processing, computer vision, bioinformatics and more generally areas involving complex datatypes. Objects of interest in such areas are usually simplified and each represented as a bag of smaller features. The occurrence frequencies of each of these features in the considered object can then be represented as a histogram. For instance, the representation of images as histograms of pixel colors, SIFT or GIST features~\citep{790410,oliva2001modeling,DJSAS09}; texts as bags-of-words or topic allocations~\citep{joachims:2002a,blei2003latent,blei2009topic}; sequences as $n$-grams counts~\citep{leslie02spectrum} and graphs as histograms of subgraphs~\citep{KasTsuIno03} all follow this principle. 

Various distances have been proposed in the statistics and machine learning literatures to compare two histograms~\citep[\S14]{deza2009encyclopedia},~\citep{rachev1991probability}. Our focus is in this paper is on the family of transportation distances, which is both well motivated theoretically~\citep[\S7]{villani},~\citep[\S5]{rachev1991probability} and works well empirically~\citep{rubner1997earth,RubTomGui00,Pele-iccv2009}. Transportation distances are particularly popular in computer vision, where, after the influential work of~\citet{rubner1997earth}, they were called \emph{Earth Mover's Distances} (EMD).

Transportation distances in machine learning can be thought of as meta-distances that build upon a \emph{metric on the features} to form a \emph{distance on histograms of features}. Such a metric, which is known in the computer vision literature as the \emph{ground metric}\footnote{Since the terms \emph{metric} and \emph{distance} are interchangeable mathematically speaking, we will always use the term \emph{metric} for a metric between features and the term \emph{distance} for the resulting transportation distance between histograms, or more generally any other distance on histograms.}, is the unique parameter of transportation distances. In their seminal paper,~\citet{RubTomGui00} argue that, ``in general, the ground distance can be any distance and will be chosen according to the problem at hand''. To our knowledge, the ground metric has always been considered a priori in all applications of EMD in machine learning. To be more precise, EMD has only been applied to datasets where such a metric was available and motivated by prior knowledge. We argue that this is problematic in two senses: first, this restriction limits the application of transportation distances to problems where such a knowledge exists. Second, even when such an a priori knowledge is available, we argue that there cannot be a ``universal'' ground metric that will be suitable for all learning problems involving histograms on such features. As with all parameters in machine learning algorithms, the ground metric should be selected adaptively. Our goal in this paper is to propose \emph{ground metric learning} algorithms to do so. 

This paper is organized as follows: After providing some background on transportation distances in Section~\ref{sec:monge}, we propose in Section~\ref{sec:criterion} a criterion -- a difference of convex function -- to select a ground metric given a training set of histograms and a similarity measure between these histograms. We then show how to obtain local minima for that criterion using a subgradient descent algorithm in Section~\ref{sec:optimization}. We propose different starting points to initialize this descent in Section~\ref{sec:initial}. We provide a review of other relevant distances and metric learning techniques in Section~\ref{sec:rel}, in particular Mahalanobis metric learning techniques~\citep{xing2003distance,weinberger2006distance,weinberger2009distance,davis2007information} which have inspired much of this work. We provide empirical evidence in Section~\ref{sec:exp} that the distances proposed in this paper compare favorably to competing techniques. We conclude this paper in Section~\ref{sec:conclusion} by providing a few research avenues.

\subsection*{Notations} We use upper case letters $A,B,\dots$ for $d\times d$ matrices. Bold upper case letters $\bA,\bB,\dots $ are used for larger matrices; lower case letters $r,c,\dots$ are used for scalar numbers or vectors of $\RR^d$.  An upper case letter $M$ and its bold lower case $\bm$ stand for the same matrix written in $d\times d$ matrix form or $d^2$ vector form by stacking successively all its column vectors from the left-most on the top to the right-most at the bottom. The notations $\overline{\bm}$ and $\underline{\bm}$ stand respectively for the strict upper and lower triangular parts of $M$ expressed as vectors of size $\binom{d}{2}$. The order in which these elements are enumerated must be coherent in the sense that the upper triangular part of $M^T$ expressed as a vector must be equal to $\underline{\bm}$. Finally we use the Frobenius dot-product for both matrix and vector representations, written as $\dotprod{A}{B}\defeq \tr(A^TB)=\ba^T\bb$.

\section{Optimal Transportation Between Histograms}\label{sec:monge}
We recall in this section a few basic facts about mass transportation for two histograms. A more general and technical introduction is provided by~\citet[Introduction \& \S7]{villani}; practical insights and motivation for its application in machine learning can be found in~\citet{RubTomGui00}; a recent review of different extensions and particular cases of EMD can be found in~\cite[\S2]{Pele-iccv2009}.
\subsection{Transportation Polytopes}\label{subsec:tptd} For two scalar histograms $r$ and $c$ of sum $1$ and dimension $d$, represented in the following as column vectors $r=(r_1,\ldots,r_d)^T$ and
$c=(c_1,\ldots,c_d)^T$ of the canonical simplex $\cs=\{u\in\RR^d_+\;|\;\norm{u}_1=1\}$ of dimension $d-1$, the polytope $U(r,c)$ of transportation plans that map $r$ to $c$ is the set of $d\times d$ matrices with coefficients in $\RR^+$ such that their row and columns marginals are equal to $r$ and $c$ respectively, that is, writing $\one_d\in\RR^d$ for the column vector of ones,
$$
U(r,c) = \{ F\in\RR_+^{d\times d}\; |\; F\one_d = r,\, F^\top\one_d = c \}.
$$
$U(r,c)$ is a polytope of dimension $d^2-2d+1$ in the general case where $r$ and $c$ have positive coordinates. $U(r,c)$ is also known in the operations research and statistical literatures as the set of transportation plans~\citep{rachev1998mass} and contingency tables or two-way tables with fixed margins~\citep{diaconisefron}. Given two histograms $r$ and $c$, we define the following function of a $d\times d$ real matrix $A$: 
\begin{equation}\label{eq:degG}
G_{rc}(A) \defeq \min_{X\in U(r,c)} \dotprod{A}{X}.
\end{equation}
Equation \eqref{eq:degG} describes a linear program whose feasible set is defined by $r$ and $c$ and whose cost is parameterized by $A$. $G_{rc}$ is a positive homogeneous function, that is $G_{rc}(t A)=t G_{rc}(A)$ for $t\geq0$. When $M$ is a matrix taken in the pointed, convex and polyhedral cone of metric matrices,
$$
\Mcal \defeq \left\{M\in\RR^{d\times d} :\, \forall\, 1\leq i,j,k\leq d, M_{ii}=0, M_{ij}=M_{ji}, M_{ij}\leq M_{ik} + M_{kj}\right\}\subset\RR_+^{d\times d},
$$
the quantity $G_{rc}(M)$ is known as the Kantorovich-Rubinstein distance~\citep[\S7]{villani} between $r$ and $c$. To highlight the fact that $G_{rc}(M)$ can also be seen as a the evaluation of a function of $r$ and $c$ parameterized by $M$, we will use the notation $d_{M}(r,c)\defeq G_{rc}(M)$.

\subsection{Transportation Distances} The function $d_M:\Sigma_{d-1}\times\Sigma_{d-1}\rightarrow \RR$ parameterized by $M$ has the following properties: since $M$ has a null diagonal $d_{M}(r,r)$ is always zero; by nonnegativity of $M$, $d_M(r,c)\geq 0$; by symmetry of $M$, $d_{M}$ is itself a symmetric function in its two arguments. More generally, $d_M$ is a distance between histograms whenever $M$ is itself a metric, namely whenever $M\in\Mcal$~\citep[Theo. 7.3]{villani}. The distance $d_M$ bears many names and has many variations: 1-Wasserstein, Monge-Kantorovich, Mallow's~\citep{mallows1972note,levina2001earth}, Earth Mover's~\citep{RubTomGui00} in vision applications.~\citet{RubTomGui00} and more recently~\citet{Pele-iccv2009} have also proposed to extend the transportation distance to compare un-normalized histograms. Simply put, these extensions compute a distance between two unnormalized histograms $u$ and $v$ by combining any difference in the total mass of $u$ and $v$ with the optimal transportation plan that can carry the whole mass of $u$ onto $v$ if $\norm{u}_1\leq \norm{v}_1$ or $v$ onto $u$ if $\norm{v}_1\leq \norm{u}_1$. We will not consider such extensions in this work; we believe however that the approaches proposed later in this paper can be extended to handle EMD distances for unnormalized histograms. 

$d_M$ can be computed as the solution of the following Linear Program (LP), 
\begin{equation}\label{eq:lp}\tag{P0}
\BA{lll}
d_M(r,c)= &\text{minimize} & \bm^T \bx\\
&\text{subject to} & \bA \bx = \begin{bmatrix}r\\ c\end{bmatrix}_* \\
&& \bx \geq 0
\EA
\end{equation} 
where $\bA$ is the $(2d-1)\times d^2$ matrix that encodes the row-sum and column-sum constraints for $X$ to be in $U(r,c)$ as
$$\bA= \begin{bmatrix}\one_{1\times d} \otimes I_d \\ I_d \otimes \one_{1\times d} \end{bmatrix}_*,$$
$\otimes$ is Kronecker's product and the lower subscript $\begin{bmatrix}\cdot\end{bmatrix}_*$ in a matrix (resp. a vector) means that its last line (resp. element) has been removed. This modification is carried out to make sure that all constraints described by $\bA$ are independent, or equivalently that $\bA^T$ is not rank deficient. This LP can be solved using the network simplex~\citep{ford1962flows} or through more specialized network flow algorithms~\cite[\S9]{ahuja1993network}.

\subsection{Properties of $G_{rc}$} 
Because its feasible set $U(r,c)$ is a bounded polytope and its objective is linear, Problem~\eqref{eq:lp} has an optimal solution in the finite set $\ExP$ of extreme points of $U(r,c)$. $G_{rc}$ is thus the minimum of a finite set of linear functions and is by extension piecewise linear and concave~\citep[\S3.2.3]{Boyd:1072}. Its gradient is equal to $\nabla G_{rc}=\bx^\star$ whenever an optimal solution $\bx^\star$ to Problem~\eqref{eq:lp} is unique~\citep[\S5.4]{bertsimas1997introduction}. More generally and regardless of the uniqueness of $\bx^\star$, any optimal solution $\bx^\star$ of Problem~\eqref{eq:lp} is in the sub-differential $\partial G_{rc}(M)$ of $G_{rc}$ at $M$~\citep[Lem.11.4]{bertsimas1997introduction}. We use this property later in Section~\ref{sec:optimization} when we optimize the criteria considered in the section below.

\section{Criteria for Ground Metric Learning}\label{sec:criterion}
We define in this section a family of criteria to quantify the relevance of a ground metric for a given task, using a training dataset of histograms with additional information.
\subsection{Training Set: Histograms and Side Information}\label{subsec:trainingsets} Suppose now that we are given a family $(r_1,\cdots,r_n)$ of histograms in the canonical simplex with a corresponding similarity matrix $[\omega_{ij}]_{1\leq i,j\leq n}\in\RR^{n\times n}$ which quantifies how similar $r_i$ and $r_j$ are: $\omega_{ij}$ is large and positive whenever $r_i$ and $r_j$ describe similar objects and small and negative for dissimilar objects. We assume that this similarity is symmetric, $\omega_{ij}=\omega_{ji}$. The similarity of an object with itself will not be considered in the following, so we simply set $\omega_{ii}=0$ for $1\leq i\leq n$.

Let us give more intuition on how these weights $\omega_{ij}$ may be set in practice. In the most simple case, these weights may reflect a class taxonomy and be set to $\omega_{ij}>0$ whenever $r_i$ and $r_j$ come from the same class and $\omega_{ij}<0$ for two different classes. This is the setting we consider in our experiments later in this paper. Such weights may be also inferred from a hierarchical taxonomy: each weight $\omega_{ij}$ corresponding to two histograms could for instance reflect how close the respective classes of these histograms lie in the tree of classes. 

Let us introduce more notations before moving on to the next section. Since $\omega_{ij}=\omega_{ji}$ and $G_{r_ir_j}=G_{r_jr_i}$ we restrict the set of pairs of indices $(i,j)$ to  
$$\Ical \defeq \{(i,j)\,|\;i,j\in\{1,\cdots,n\}, i<j\}.$$ We also introduce two subsets of $\Ical$: 
$$\Ecal_+\defeq\{(i,j)\in\Ical\,|\;\omega_{ij}> 0\};\quad \Ecal_-\defeq\{(i,j)\in\Ical\,|\;\omega_{ij}< 0\},$$
the subsets of similar and dissimilar histograms. Finally, we write $G_{ij}$ for the $G_{r_i r_j}$ functions.

\subsection{A Local Criterion to Select the Ground Metric}\label{subsec:local} We propose to formulate the ground metric learning problem as finding a metric $M\in\Mcal$ such that the transportation distance $d_M$ induced by this metric agrees with the weights $\omega$. More precisely, this criterion will favor metrics for which, for a given pair of similar histograms $r_i$ and $r_j$ (namely $\omega_{ij}>0$), the resulting distance $G_{ij}(M)$ is small. Conversely, for a given pair of dissimilar histograms $r_i$ and $r_j$ (namely $\omega_{ij}<0$), the resulting distance $G_{ij}(M)$ should be large. The criterion should balance these two requirements, and in particular favor ground metrics for which these two ideas hold for pairs $(i,j)$ such that $\abs{\omega_{ij}}$ is large.
 
From a formal perspective, any criterion to select $M$ should consider the family of $\binom{n}{2}$ pairs
$$
\{\,\left(\omega_{11},G_{11}(M)\right),\,\cdots,\left(\omega_{n-1\,n},G_{n-1\,n}(M)\right)\,\}.
$$
Since the ordering of the histograms should not influence the criterion, only symmetric functions $(\RR\times \Sigma_{d-1})^{\binom{n}{2}}\rightarrow \RR$ of the couples of variables above should be considered. We propose in this paper a family of simple criteria: the \emph{average} value of $\omega_{ij}d_M(r_i,r_j)$,
$$
C_\infty(M) \defeq 2\sum_{(i,j)\in\Ical} \omega_{ij} G_{ij}(M),
$$
and a restriction of such an average to neighboring points,
$$
C_k(M) \defeq \sum_{i=1}^n S^+_{ik}(M) + S^-_{ik}(M),
$$
where for each index $i$, the weighted sums of distances of its \emph{similar} and \emph{dissimilar} neighbors are considered respectively in
\begin{equation}\label{eq:sikpsikm}
	S^+_{ik}(M)\defeq \sum_{j\in N^+_{ik}} \omega_{i j} G_{ij}(M),\;\text{ and }\; S^-_{ik}(M) \defeq \sum_{j\in N^-_{ik}} \omega_{i j} G_{ij}(M).
\end{equation}
These sums are computed using the sets $N^+_{ik}$ and $N^-_{ik}$, which stand for the indices of any $k$ nearest neighbours of $r_i$ using distance $d_M$, not necessarily unique, and whose indices are taken respectively in the subsets $\Ecal_{i+}\defeq \{j | (i,j) \text{ or } (j,i)\in\Ecal_+ \}$ and $\Ecal_{i-}\defeq\{j | (i,j) \text{ or } (j,i)\in\Ecal_- \}$. 
We adopt the convention that $N^+_{ik}=\Ecal_{i+}$ whenever $k$ is larger than the cardinality of $\Ecal_{i+}$, and follow the same convention for $N^-_{ik}$. This convention makes our notation $C_\infty$ consistent with the definition of $C_k$ since one can indeed check that $C_k=C_\infty$ for $k$ large enough, and notably for $k\geq n$. 
Since the techniques we propose below apply to both cases where $k$ is finite or infinite, we only consider in the following an extended index $k\in[1,\infty]$ and its corresponding criterion $C_k$.

\subsection{Metrics of Interest} Because $C_k$ is positively homogeneous, the problem of minimizing $C_k$ over the pointed cone $\Mcal$ is either unbounded or trivially solved for $M=0$. To get around this issue, we restrict our search to the intersection of $\Mcal$ and the unit sphere in $\RR^{d\times d}$ of a suitable matrix norm, that is 
\begin{equation}\label{eq:mcal1}
\Mcal_1 = \Mcal \cap B_1,
\end{equation}
where $B_1=\{A\in\RR^{d\times d}\; | \;\norm{A}_\cdot \leq 1\}$ is defined by an arbitrary matrix norm $\norm{A}_\cdot$ such that the unit ball $B_1$ is convex. Using criteria $C_k$, learning a ground metric from the family of points $(r_1,\cdots,r_n)$ and weights $(\omega_{ij})_{(i,j)\in\Ical}$ boils down to finding an optimal solution to problem
\begin{equation}\label{eq:mainprob}\tag{P1}
\min_{M\in\Mcal_1} C_k(M).
\end{equation}
Problem~\eqref{eq:mainprob} has $\binom{d}{2}$ variables, one for each upper-diagonal term in $M$. The feasible set $\Mcal_1$ is closed, convex, bounded and $C_k$ is piecewise linear. As a consequence, Problem~\eqref{eq:mainprob} admits at least one optimal solution.

\section{$C_k$ as a Difference of Convex Functions}\label{sec:optimization}

Since each function $G_{ij}$ is concave, $C_k$ can be cast as a Difference of Convex (DC) functions~\citep{horst1999dc}:
$$
C_k(M) \defeq S^-_{k}(M) -\,  \minus S^+_{k}(M)
$$
where both $S^-_{k}(M)\defeq \sum_{i=1}^n S^-_{ik}(M)$ and $\minus S^+_{k}(M)\defeq \sum_{i=1}^n \minus S^+_{ik}(M)$ are convex, by virtue of the convexity of each of the terms $S^-_{ik}$ and $\minus S^+_{ik}$ defined in Equation~\eqref{eq:sikpsikm}. This follows from the concavity of each function $G_{ij}$ and the fact that such functions are weighted by negative factors, $\omega_{ij}$ for $(i,j)\in\Ecal_-$ and $\minus\omega_{ij}$ for $(i,j)\in\Ecal_+$. We propose in this section an algorithm to obtain a local minimizer of Problem~\eqref{eq:mainprob} that takes advantage of this decomposition.

\subsection{Subdifferentiability of $C_k$} The gradient of $C_k$ computed at a given metric matrix $M$ is
$$\nabla C_k(M) = \nabla S^-_{k}(M) -\,  \minus\nabla S^+_{k}(M),$$ where 
$$\nabla S^{-}_{k}(M) = \sum_{i=1}^n\sum_{j\in N^{+}_{ik}} \omega_{i j} X_{ij}^\star,\quad \minus\nabla S^{+}_{k}(M) = \sum_{i=1}^n \sum_{j\in N^{-}_{ik}} \minus\omega_{i j} X_{ij}^\star,$$
whenever all solutions $X_{ij}^\star$ to the linear programs $G_{ij}$ considered in $C_k$ are unique and whenever the set of $k$ nearest neighbors of each histogram $r_i$ is unique. More generally, by virtue of the property that any optimal solution $X_{ij}^\star$ is in the sub-differential $\partial G_{ij}(M)$ of $G_{ij}$ at $M$ we have that 
$$\sum_{i=1}^n\sum_{j\in N^+_{ik}} \omega_{i j} X_{ij}^\star \in \partial S^-_{k}(M),\quad \sum_{i=1}^n\sum_{j\in N^+_{ik}}\minus\omega_{ij} X_{ij}^\star \in \partial \minus S^+_{k}(M),$$ 
regardless of the unicity of the $k$ nearest neighbors of each histogram $r_i$. The details of the computation of $S^-_{k}(M)$ and one of its subgradients are given in Algorithm~\eqref{algo:sub}. 
The computations for $S^+_{k}(M)$ follow the same route; we use the abbreviation $S^{\{+,-\}}_{k}(M)$ to consider either of these two cases in our algorithm outline.

\begin{algorithm}
\begin{algorithmic}
\caption{Subgradient and Objective Computation for $S_k^{\{+,-\}}(M)$\label{algo:sub}}  
\STATE \textbf{Input}:  $M \in \Mcal_1$. \textbf{optional}: warm starts $\{\tilde{X}_{ij}\}$.
\FOR {$(i,j)\in \Ecal_{\{+,-\}}$}
\STATE Compute the optimum $z_{ij}^\star$ and an optimal solution $X_{ij}^\star$ for Problem~\eqref{eq:lp}, using the network simplex for instance, with cost vector $\bm$ and constraint vector $[r_i; r_j]_*$. \textbf{optional}: use warm starts $\{\tilde{X}_{ij}\}$.
\ENDFOR
\STATE Set $G=0, z=0$.
\FOR {$i\in \{1,\cdots,n\}$}
\STATE Compute the neighborhood set $N_{ik}^{\{+,-\}}$ by ranking all $z_{ij}^\star$ in $\Ecal_{i\{+,-\}}$.
\FOR {$j\in N_{ik}^{\{+,-\}}$}
\STATE $G \gets G + \omega_{ij} X^\star_{ij}$. 
\STATE $z\gets z+ \omega_{ij} z^\star$.
\ENDFOR
\ENDFOR
\STATE \textbf{Output} $z$ and $\nabla = \overline{\bg} + \underline{\bg}$; \textbf{optional}: return current solutions $\{X_{ij}^\star\}$ as warm starts for the next iteration.
\end{algorithmic}
\end{algorithm}

\subsection{Localized Linearization of the Concave Part of $C_k$}  We describe in Algorithm~\eqref{algo:optim} a simple approach to minimize $C_k$ locally based on a projected subgradient descent and a local linearization of the concave part of $C_k$. Algorithm~\eqref{algo:optim} runs a subgradient descent on $C_k$ using two nested loops. In the first loop parameterized with variable $p$, $S^+_k$ (the concave part of $C_k$) and a point $\nabla_{+}$ in its subdifferential are computed using the current metric $M_p$. Using this value and the subgradient $\nabla_{+}$, the concave part $S^+_k$ of $C_k$ can be locally approximated by its first order Taylor expansion,
$$C_k(M)\approx S^-_k(M) + S^+_k(M_p) + \nabla_{+}^T (M-M_p).$$
This approximation is convex, larger than $C_k$ and can be minimized in an inner loop using a projected gradient descent. When this convex function has been minimized up to a sufficient precision, we obtain a point
$$M_{p+1} \in \argmin_{M\in\Mcal_1} S^-_k(M) + S^+_k(M_p) + \nabla_{+}^T (M-M_p).$$
We increment $p$ repeat the step described above. The algorithm terminates when sufficient progress in the outer loop has been realized, at which point either the matrix computed in the last iteration, or that for which the objective has been minimal so far, is returned as the output of the algorithm.

Algorithm~\eqref{algo:optim} fits the description of simplified DC algorithms~\citep[\S4.2]{tao1997convex} to minimize a difference of convex functions $g-h$ in the case where either $g$ or $h$ is a convex polyhedral function. In this paper \emph{both} functions $S_k^-(M)$ and $\minus S_k^+(M)$ are convex polyhedral;  The overall quality of this local minima is directly linked to the quality of the initial point $M_0$. Choosing a good $M_0$ is thus a crucial factor of our approach. We provide a few options to define $M_0$ in Section~\ref{sec:initial}.

\begin{algorithm}
\begin{algorithmic}
\caption{Projected Subgradient Descent to minimize $C_k$\label{algo:optim}}
\STATE \textbf{Input} $M_0 \in \Mcal_1$ (see Section~\ref{sec:initial}), gradient step $t_0$.
\STATE $t\gets 1$.
\STATE $p\gets 0,\quad M^{\text{out}}_0\gets M_0$.
\REPEAT 
\STATE Compute $\nabla_+$ and $z_+$ of $S_k^+$ using Algorithm~\eqref{algo:sub} with $M^{\text{out}}_p$.
\STATE $q\gets 0,\quad M^{\text{in}}_0\gets M^{\text{out}}_p$.
\REPEAT
\STATE Compute $\nabla_-$ and $z_-$ of $S_k^-$ using Algorithm~\eqref{algo:sub} with $M^{\text{in}}_q$ and warm-starts $\tilde{X}_{ij}, (i,j)\in\Ecal^-$ if defined; Set $\tilde{X}_{ij}\gets X^\star_{ij}$ for $(i,j)\in\Ecal^-$.
\STATE If $q=0$, set $z^{\text{out}}_p \gets z^{\text{in}}_q$.
\STATE Set $z^{\text{in}}_q \gets z_- + z_+ + \nabla_+^T(\underline{\bm}^{\text{in}}_q-\underline{\bm}^{\text{out}}_p)$
\STATE Set $M^{\text{in}}_{q+1}\gets P_{\Mcal_1}\left(\underline{\bm}^{\text{in}}_q-\frac{t_0}{\sqrt{t}}(\nabla_+ +\nabla_-) \right)$.
\STATE $q\gets q+1$.
\STATE $t\gets t+1$.
\UNTIL {$q<q_{\max}$ or insufficient progress for $z^{\text{in}}_q$}.
\STATE $M^{\text{out}}_{p+1}\gets M^{\text{in}}_{q}$.
\STATE $p\gets p+1$.
\UNTIL {$p<p_{\max}$ or insufficient progress for $z^{\text{out}}_p$}.
\STATE \textbf{Output} $\bd_p$ on $\Mcal_p$.
\end{algorithmic}
\end{algorithm}

\section{Initial Points}\label{sec:initial}

Algorithm~\eqref{algo:optim} converges to a local minima of $C_k$ in $\Mcal_1$. We argue that this local solution can only provide a good approximation of the global minima if the initial point $M_0$ itself is already a good initial guess, not too far from the global optimum of $C_k$. 

\subsection{The $l_1$ Distance as a Transportation Distance} The $l_1$ distance between histograms can provide an educated guess to define an initial point $M_0$ to optimize $C_k$. Indeed, the $l_1$ distance can be interpreted as the Kantorovich-Rubinstein distance\footnote{Rigorously speaking, both distances are equal up to a factor 2, that is $\frac{1}{2}\|r-c\|_1 = d_{M_{\one}}(r,c)$} seeded with the uniform ground metric $M_{\one}$ defined as $M_{\one}(i,j)=\one_{i=j}$. Because the $l_1$ distance is itself a popular distance to compare histograms, we consider $M_\one$ in our experiments to initialize Algorithm~\eqref{algo:optim}. This starting point does not, however, exploit the information provided by the histograms $\{r_i, 1\leq i \leq n\}$ and weights $\{\omega_{ij}, (i,j)\in\Ical\}$. In order to do so, we approximate $C_k$ by a linear function of $M$ in Section~\ref{subsec:linapprox}, and show that a minimizer of this approximation can provide a better way of setting $M_0$, as shown later in the experimental section.

\subsection{Linear Approximations to $C_k$}\label{subsec:linapprox}
We propose to form an initial point $M_0$ by replacing the optimization underlying the computation of each distance $G_{ij}(M)$ by a dot product,
\begin{equation}\label{eq:linapprox}
G_{ij}(M) = \min_{X\in U(r_i,r_j)} \dotprod{M}{X} \approx \dotprod{M}{\Xi_{ij}}
\end{equation}
where $\Xi_{ij}$ is a $d\times d$ matrix. We discuss several choices to define matrices $\Xi_{ij}$ later in Section~\ref{sub:typical}. We use these approximations to define the criteria
\begin{equation}\label{eq:infty}
\chi_\infty(M) = 2 \sum_{(i,j)\in\Ical} \omega_{ij} \dotprod{M}{\Xi_{ij}} = 2\dotprod{M}{\sum_{(i,j)\in\Ical} \omega_{ij} \Xi_{ij}}
\end{equation}
in the case where $k=\infty$, or 
\begin{multline}\label{eq:uptok}
\chi_k(M) = \sum_{i=1}^n \sum_{j\in N^-_{ik}(M_{\one})}\dotprod{M}{\Xi_{ij}} + \sum_{j\in N^+_{ik}(M_{\one})}\dotprod{M}{\Xi_{ij}}\\= \dotprod{M}{\sum_{i=1}^n\sum_{j\in N^-_{ik}(M_{\one})}\omega_{ij}  \Xi_{ij} + \sum_{j\in N^+_{ik}(M_{\one})} \omega_{ij} \Xi_{ij}}
\end{multline}
when $k<\infty$. Note that when $k$ is finite, and only in that case, the $k$ nearest neighbors of each histogram $r_i$ need to be selected first with a metric; we use $M_{\one}$ for this purpose. Although this trick may not be satisfactory, we observe that similar approaches have been used by~\citet{weinberger2009distance} to seed their algorithms with near neighbors in the initial phase of their optimization. Note that such a trick is not needed when $k=\infty$ which, in practice, seems to yield better results as explained later in the experimental section. In both cases where $k=\infty$ and $k<\infty$, $\chi_k$ is a linear function of $M$ which, for our purpose, needs to be minimized over $\Mcal_1$:

\begin{equation}\label{eq:lininmcal}\tag{P2}
\min_{M\in\Mcal_1} \chi_k(M) = \min_{M\in\Mcal_1} \dotprod{M}{\Xi_k} 
\end{equation}
where $\Xi_k$ is a $d\times d$ matrix equal to the relevant sums in the right hand sides of either Equation~\eqref{eq:infty} or~\eqref{eq:uptok} depending on the value of $k$. This problem has a linear objective and a convex feasible set. If the norm defining the unit ball $B_1$ in Equation~\eqref{eq:mcal1} is the $l_1$ norm, that is the sum of the absolute values of all coefficients in a matrix, then Problem~\eqref{eq:lininmcal} is a linear program with $O(d^3)$ constraints. For large $d$, this formulation might be intractable. We propose to consider instead the $l_2$ norm unit ball for $B_1$ which yields an alternative form for Problem~\eqref{eq:mainprob}, where the constraint $M\in B_1$ is replaced by a regularization term 
\begin{equation}\label{eq:metricnearness1}\tag{P3}
\min_{M\in\Mcal} \lambda\dotprod{M}{\Xi_k} + \norm{M}^2_2 = \min_{M\in\Mcal} \norm{M + \frac{\lambda}{2} \Xi_k}_2^2, \quad \lambda>0 
\end{equation}
\citet[Algorithm 3.1]{brickell2008metric} have proposed recently a triangle fixing algorithm to solve problems of the form 
\begin{equation}\label{eq:metricnearness2}\tag{P4}
\min_{M\in\Mcal} \norm{M - H}_2,
\end{equation}
where $H$ is a pseudo-distance, that is a \emph{symmetric}, \emph{zero on the diagonal} and \emph{nonnegative} matrix. It is however straightforward to check that each of these three conditions, although intuitive when considering the metric nearness problem as defined in~\citep[\S2]{brickell2008metric}, are not necessary for Algorithm (3.1) in~\citep[\S3]{brickell2008metric} to converge. This algorithm is not only valid for non-symmetric matrices $H$ as pointed out by the authors themselves, but it is also applicable to matrices $H$ with negative entries and non-zero diagonal entries. Problem~\eqref{eq:metricnearness1} can thus be solved by replacing $H$ by $-\frac{\lambda}{2}\Xi_k$ in Problem~\eqref{eq:metricnearness2} regardless of the sign of the entries of $\Xi$.

We conclude this section by mentioning that other approaches can be considered to minimize the dot product $\dotprod{M}{\Xi}$ using alternative norms and methods.~\citet{frangioni2005new} propose for instance to handle linear programs in the intersection between the cone of distances and the set of polyhedral constraints $\{M_{ik}+M_{ki}+M_{ij}\leq 2\}$ which defines what is known as the metric polytope. These approaches are however more involved computationally and we leave such extensions for future work.
 
\subsection{Representative Tables}\label{sub:typical}
The techniques presented in Section~\ref{subsec:linapprox} above build upon a linear approximation of each function $G_{ij}(M)$ as $\dotprod{M}{\Xi_{ij}}$ by selecting a particular matrix $\Xi_{ij}$ such that $G_{ij}(M)\approx\dotprod{M}{\Xi_{ij}}$. We propose in this section to obtain such an approximation by considering an arbitrary and representative transportation table in $U(r,c)$. More precisely, we propose to use a simple proxy for the optimal transportation distance: the dot-product of $M$ with a matrix that lies at the center of $U(r,c)$.

\subsubsection{Independence Table}
Many candidate tables can qualify as valid centers of general polytopes as discussed for instance in~\citep[\S8.5]{Boyd:1072}. There is, however, a particular table in $U(r,c)$ which is easy to compute and which has been considered as a central point of $U(r,c)$ in previous work: the \emph{independence} table $rc^T$~\citep{good1963maximum}. The table $rc^T$, which is trivially in $U(r,c)$ because $rc^T\one_d=r$ and $cr^T\one_d=c$, is also the maximal entropy table in $U(r,c)$, that is the table which maximizes 
\begin{equation}
h(X)\defeq -\sum_{p,q=1}^d X_{pq} \log X_{pq}.
\end{equation} 
Using the independence table to approximate $G_{ij}$, that is using the approximation 
$$\min_{F\in U(r_i,r_j)} \dotprod{M}{F} \approx r_i^T M r_j,$$
yields the averages independence tables,
\begin{equation}\label{eq:indepedentinf}
\Xi_{k}= \begin{cases} \sum_{ij} \omega_{ij} r_i r_j^T, &\text{ if } k=\infty.\\
\sum_{i=1}^n\sum_{j\in N^-_{ik}(M_{\one})}\omega_{ij} r_ir_j^T + \sum_{j\in N^+_{ik}(M_{\one})} \omega_{ij}  r_ir_j^T, & \text{ if } k<\infty.
\end{cases}
\end{equation}
Note however that this approximation tends to overestimate substantially the distance between two similar histograms. Indeed, it is easy to check that $r^TM r$ is positive whenever $M$ is a definite distance and $r$ has positive entropy. In the case where all coordinates of $r$ are equal to $1/d$, $r^T M r$ is simply $\norm{M}_1/d^2$.

\subsubsection{Typical Table}\label{subsub:barv}
~\citet{barvinok2010does} argues more recently that most transportation tables are close to the so-called \emph{typical} table $T$ of the transportation polytope and not, as was hinted by~\citet{good1963maximum}, to the independence table. We briefly explain the concentration result obtained in~\citep[\S1.5]{barvinok2010does}.~\citeauthor{barvinok2010does} proves that, under the condition that $r$ and $c$ do not have too small coefficients, for any table $X$ sampled uniformly on $U(r,c)$ the difference between the sum of a subset of coefficients of $X$ and the sum of the same coefficients in the typical table $T$ is small with high-probability. Writing $S\subset \{1,\cdots,n\}^2$ for a set of indices, and $\sigma_S(X)=\sum_{p,q\in S}X_{pq}$ for the corresponding sum of coefficients, we have that for sets $S$ big enough,
$$
P\left\{X\in U(r,c), (1-\ve) \sigma_S(T) \leq \sigma_S(X) \leq (1+\ve) \sigma_S(T) )\right\} \geq 1-2de^{-\kappa d}
$$
where $\kappa$ and $\ve$ depends on the \emph{smoothness} of $r$ and $c$, that is the magnitude of their smallest coefficients. The typical table $T_{ij}$ of two histograms $r_i$ and $r_j$ is defined~\citep[\S1.2]{barvinok2010does} as the table in the polytope $U(r_i,r_j)$ which maximizes the concave function $g:\RR_{++}^{d\times d}\rightarrow\RR$
\begin{equation}\label{eq:typicalmin}
g(X)= \sum_{p,q=1}^d (X_{pq}+1)\ln(X_{pq}+1) - X_{pq}\ln(X_{pq}).
\end{equation}
Computing the typical table directly is not computationally tractable for large values of $d$.~\citet[p.350]{barvinok2009asymptotic} provides however a different characterization of $T_{ij}$ as
$$T_{ij}=\left[\frac{e^{-u_p-v_q}}{1-e^{-u_p-v_q}}\right]_{p,q\leq d},$$
where the vectors $u$ and $v$ in $\RR^d$ are the unique minimizers of the convex program 
\begin{equation}\label{eq:typical}\tag{P5}
\min_{u,v> 0} r_i^Tu + r_j^Tv - \sum_{p,q=1}^d \log\left(1-e^{-u_p-v_q}\right).
\end{equation}
Both gradient and Hessian of the objective of Problem~\eqref{eq:typical} have a simple form. The Hessian can be expressed in a block form where the diagonal blocks are themselves diagonal matrices and the off-diagonal blocks are of rank $1$. $u$ and $v$ can thus be easily computed using second-order methods. The resulting matrices $\Xi_k$ are thus
\begin{equation}\label{eq:typicalsums}
\Xi_{k}= \begin{cases} \sum_{ij} \omega_{ij} T_{ij}, &\text{ if } k=\infty.\\
\sum_{i=1}^n\sum_{j\in N^-_{ik}(M_{\one})}\omega_{ij} T_{ij} + \sum_{j\in N^+_{ik}(M_{\one})} \omega_{ij}  T_{ij}, & \text{ if } k<\infty.
\end{cases}
\end{equation}
We also note that, as for the independence table, $\dotprod{M}{T_{ij}}$ will be significantly larger than $0$ when $r_i$ and $r_j$ are similar.

Let us provide some intuition on the idea behind using the average typical table in Problem~\eqref{eq:metricnearness1}. The solution to Problem~\eqref{eq:metricnearness1} will be a metric $M$ which will have high coefficients $M_{pq}$ for any pair of features $1\leq p,q\leq d$ such that $\Xi_{pq}$ is negative, namely a pair $(p,q)$ such that the value of $X_{pq}$ of a transportation plan $X$ sampled uniformly in each $U(r_i,r_j)$ is \emph{typically} high on average for all \emph{mismatched} histograms pairs. $M_{pq}$ will be on the contrary small for a pair of features $(p,q)$ such that the value of $X_{pq}$ is \emph{typically} high in transportations plans between \emph{similar} histograms.


To recapitulate the results of this section, we propose to approximate $C_k$ by a linear function and compute its minimum in the intersection of the unit ball and the cone of matrices. This linear objective can be efficiently minimized using a set of tools proposed by~\citep{brickell2008metric} adapted to our problem. In order to do so, the unit ball considered to define the feasible set $\Mcal_1$ must be the unit ball of the Frobenius norm of matrices. In order to propose such an approximation, we have used the \emph{independence} and \emph{typical} tables as representative points of the polytopes $U(r,c)$. The successive steps of the computations that yield an initial point $M_0$ are spelled out in Algorithm~\eqref{algo:initial}. 

\begin{algorithm}
\begin{algorithmic}
\caption{Initial Point $M_0$ to minimize $C_k$\label{algo:initial}}  

\STATE Set $\Xi=0$.
\FOR {$i\in \{1,\cdots,n\}$}
\IF {$k=\infty$} 
\STATE set $N_{ik}^{\{+,-\}}=\Ecal_{i\{+,-\}}$.
\ELSE
\STATE Compute the neighborhood sets $N_{ik}^{+}$ and $N_{ik}^{-}$ of histogram $r_i$ using an arbitrary distance, \eg\, the $l_1$ distance. 
\ENDIF
\FOR {$j\in N_{ik}^{+} \cup N_{ik}^{-}$}
\STATE Compute a center $\Xi_{ij}$ of $U(r_i,r_j)$, \eg\; either the typical or independence table.
\STATE $\Xi \gets \Xi + \omega_{ij} \Xi_{ij}$. 
\ENDFOR
\ENDFOR
\STATE Set $M_0\gets \min_{M\in\Mcal} \norm{M +\frac{\lambda}{2} \Xi}_2$ using \citep[Algorithm 3.1]{brickell2008metric}.
\STATE \textbf{Output} $M_0$. \textbf{optional}: regularize $M_0$ by setting $M_0\gets \lambda M_0+ (1-\lambda) M_\one$.
\end{algorithmic}
\end{algorithm}


\section{Related Work}\label{sec:rel}

\subsection{Metrics on the Probability Simplex} \citet[\S14]{deza2009encyclopedia} provide an exhaustive list of metrics for probability measures, most of which apply to probability measures on $\RR$ and $\RR^d$. When narrowed down to distances for probabilities on unordered discrete sets -- the dominant case in machine learning applications --~\citet[\S2]{RubTomGui00} propose to split the most commonly used distances into two families: \emph{bin-to-bin} distances and \emph{cross-bin} distances. 

\textbf{Bin-to-bin distances} only compare the $d$ couples of bin-counts $(r_i,c_i)_{i=1..d}$ independently to form a distance between $r$ and $c$: the Jensen-divergence, $\chi_2$, Hellinger, total variation distances and more generally Csizar $f$-divergences ~\citep[\S3.2]{amar01b} all fall in this category. Notice that each of these distance is known to work usually better for histograms than a straightforward application of the Euclidean distance as illustrated for instance in~\citep[Table 4]{chapelle99SVMs} or in our experimental section. This can be explained in theory using geometric~\cite[\S3]{amar01b} or statistical arguments~\citep{aitchison2005compositional}.

Bin-to-bin distances are easy to compute and accurate enough to compare histograms when all $d$ features are sufficiently distinct. When, on the contrary, some of these features are known to be similar, either because of statistical co-occurrence (\eg\, the words \texttt{Nadal} and \texttt{Federer}) or through any other form of prior knowledge (\eg\,color or amino-acid similarity) then a simple bin-to-bin comparison may not be accurate enough as argued by~\citep[\S2.2]{RubTomGui00}. In particular, bin-to-bin distances are large between histograms with distinct supports, regardless of the fact that these two supports may in fact describe very similar features.

\textbf{Cross-bin distances} handle this issue by considering all $d^2$ possible pairs $(r_i,c_j)$ of cross-bin counts to form a distance. The most simple cross-coordinate distance for general vectors in $\RR^d$ is arguably the Mahalanobis family of distances, $$d^\Omega(x,y)=\sqrt{(x-y)^T \Omega (x-y)},$$ where $\Omega$ is a positive semidefinite $d\times d$ matrix. The Mahalanobis distance between $x$ and $y$ can be interpreted as the Euclidean distance between $Lx$ and $Ly$ where $L$ is a Cholesky factor of $\Omega$ or any square root of $\Omega$. Learning such linear maps $L$ or $\Omega$ directly using labeled information has been the subject of a substantial amount of research in recent years. We briefly review this literature in the following section.

\subsection{Mahalanobis Metric Learning}
\citet{xing2003distance}, followed by~\citet{weinberger2006distance} and~\citet{davis2007information} have proposed different algorithms to learn the parameters of a Mahalanobis distance, that is either a positive semi-definite matrix $\Omega$ or a linear map $L$. These techniques define first a criterion $C$ and a feasible set of candidate matrices to obtain, through optimization algorithms, a relevant matrix $\Omega$ or $L$. The criteria we propose in Section~\ref{sec:criterion} are modeled along these ideas.~\citet{weinberger2006distance} were the first to consider criteria that only use nearest neighbors, which inspired in this work the proposal of $C_k$ for finite values of $k$ in Section~\ref{subsec:local} as opposed to considering the average over all possible distances as in~\citep{xing2003distance} for instance.

We would like to insist at this point in the paper that Mahalanobis metric learning and ground metric learning have very little in common conceptually: Mahalanobis metric learning algorithms produce a $d\times d$ positive semidefinite matrix or a linear operator $L$. Ground metric learning produces instead a metric matrix $M$. These sets of techniques operate on very different mathematical objects.

It is also worth mentioning that although Mahalanobis distances have been designed for general vectors in $\RR^d$, and as a consequence can be applied to histograms, there is however, to our knowledge, no statistical theory which motivates their use on the probability simplex.

\subsection{Metric Learning in $\Sigma_{d-1}$}
\citet{leb06metric} has proposed to learn a bin-to-bin distance in the probability simplex using a parametric family of distances parameterized by a histogram $\lambda\in \Sigma_{d-1}$ defined as
$$
d_\lambda(r,c) = \arccos \left( \sum_{i=1}^d \sqrt{\frac{r_i \lambda_i}{r^T\lambda }} \sqrt{\frac{c_i \lambda_i}{c^T\lambda}}\right)
$$
This formula can be simplified by using the perturbation operator proposed by~\citet[p.46]{Aitchison:1986}: 
$$
\forall r,\lambda\in\Sigma_{d-1},\quad r \odot \lambda \defeq \frac{1}{r^T \lambda}(r_1\lambda_1, \cdots , r_d\lambda_d)^T
$$
\citeauthor{Aitchison:1986} argues that the perturbation operation can be naturally interpreted as an addition operation in the simplex. Using this notation, the distance $d_\lambda(r,c)$ becomes the simple Fisher metric applied to the perturbed histograms   $r\odot \lambda$ and $c \odot \lambda$,
$$
d_\lambda(r,c) = \arccos \dotprod{\sqrt {r \odot \lambda}}{\sqrt {c \odot \lambda}}.
$$
Using arguments related to the fact that the distance should vary accordingly to the density of points described in a dataset,~\citet{leb06metric} proposes to learn this perturbation $\lambda$ in a semi-supervised context, that is making only use of observed histograms but no other side-information. Because of this key distinction we do not consider this approach in the experimental section.

\section{Experiments}\label{sec:exp}
We provide in this section a few details on the practical implementation of Algorithms~\eqref{algo:sub},~\eqref{algo:optim} and~\eqref{algo:initial}. We follow by presenting empirical evidence that ground metric learning improves upon other state-of-the-art metric learning techniques when considered on normalized histograms.
\subsection{Implementation Notes}
Algorithms~\eqref{algo:sub},~\eqref{algo:optim} and~\eqref{algo:initial} were implemented using several optimization toolboxes. Algorithm~\eqref{algo:sub} requires the computation of several transportation problems. We use the \emph{CPLEX Matlab API} implementation of network flows with warm starts to that effect. The computational gains we obtain by using the API, instead of using a function call to the \emph{CPLEX matlab toolbox}  or to the \emph{Mosek} solver are approximately 4 fold. These benefits come from the fact that only the constraint vector in Problem~\eqref{eq:lp} needs to be updated at each iteration of the first loop of Algorithm~\eqref{algo:sub}. We use the \emph{metricNearness} toolbox\footnote{\texttt{http://people.kyb.tuebingen.mpg.de/suvrit/work/progs/metricn.html}} to carry out both the projections of each inner loop iteration of Algorithm~\eqref{algo:optim}, as well as the last minimization of Algorithm~\eqref{algo:initial}. We compute Typical tables using the \emph{fminunc} Matlab function, with the gradient and the Hessian of the objective of Problem~\eqref{eq:typical} provided as auxiliary functions. Since \emph{fminunc} is by definition an unconstrained solver, its solution $(u^\star,v^\star)$ is kept if both $u^\star$ and $v^\star$ satisfy positivity constraints, which is the case for a large majority of pairs of histograms. Whenever these constraints are violated we optimize again this problem by using a slower constrained Newton method.
 
\subsection{Images Classification Datasets}\label{subsec:images}   We sample randomly $2N$ classes taken in the Caltech-256 family of images and consider 70 images in each class. Each image is represented as a normalized histogram of GIST features, obtained using the LEAR implementation\footnote{\texttt{http://lear.inrialpes.fr/software}} of GIST features~\citep{DJSAS09}. These features describe $8$ edge directions at mid-resolution computed for each patch of a $4\times 4$ grid on each image. Each feature histogram is of dimension $d=8\times 4\times 4=128$ and subsequently normalized to sum to one. 

We split these classes into two sets of $N$ classes, $(a_1,\cdots,a_{N})$ and $(a_{N+1},\cdots,a_{2N})$ and study the resulting $N^2$ binary classification problems that arise from each pair of classes in $$\{a_1,\cdots,a_{N}\} \times \{a_{N+1},\cdots,a_{2N}\}.$$  
For each of these $N^2$ binary classification task we split the $70+70$ points from both classes into $30+30$ points to form a training set and $40+40$ points to form a test set.  This amounts to having $n=60$ training points following the notations introduced in Section~\ref{subsec:trainingsets}. 

\subsection{Distances used in this benchmark} 
\subsubsection{Bin-to-bin distances} We consider the $l_1, l_2$ and Hellinger distances on GIST features vectors,
$$
l_1(r,c)= \norm{r-c}_1, \quad l_2(r,c)= \norm{r-c}_2, \quad \Hcal(r,c)= \norm{\sqrt r- \sqrt c}_2, 
$$
where $\sqrt{r}$ is the vector whose coordinates are the squared root of each coordinate of $r$. 
\subsubsection{Mahalanobis distances}\label{subsub:Mah} We use the public implementations of LMNN~\citep{weinberger2009distance} and ITML~\citep{davis2007information} to learn two different Mahalanobis distances for each task. We run both algorithms with default settings, that is $k=3$ for LMNN and $k=4$ for ITML. We use these algorithms on the Hellinger representations $\{\sqrt{r_i},\,i=1,\cdots,n\}$ of all histograms originally in the training set. We have considered this representation because the Euclidean distance of two histograms using the Hellinger map corresponds exactly to the Hellinger distance~\cite[p.57]{amar01b}. Since the Mahalanobis distance builds upon the Euclidean distance, we argue that this representation is more adequate to learn Mahalanobis metrics in the probability simplex. The significant gain in performance observed in Figure~\ref{fig:lmnnanco} that is obtained through this simple transformation confirms this intuition.

\subsubsection{Ground Metric Learning} We learn ground metrics using the following settings. In each classification task, and for two images $r_i$ and $r_j$, $1\leq i\neq j\leq 60$, each weight $\omega_{ij}$ is set to $1$ if both histograms come from the same class and to $-1$ if they come from different classes. The weights $\omega_{ij}$ are further normalized to ensure that $\sum_{(i,j)\in\Ecal^+}\omega_{ij}=1$ and $\sum_{(i,j)\in\Ecal^-}\omega_{ij}=-1$. As a consequence the total sum of weights $\sum_{(i,j)\in\Ecal}\omega_{ij}$ is equal to $0$. The neighborhood parameter $k$ is set to $3$ to be directly comparable to the same parameter used for ITML and LMNN. The subgradient stepsize $t_0$ of Algorithm~\eqref{algo:optim} is set to $=0.1$, guided by preliminary experiments and by the fact that, because of the normalization of the weights $\omega_{ij}$ introduced above, both the current iteration $M_k$ in Algorithm~\eqref{algo:optim} and the gradient steps $\nabla_+$ or $\nabla_-$ all have comparable norms as matrices. We perform a minimum of $50\times 0.8^p$ gradient steps in each inner loop and set $p_{\max}$ to $8$. Each inner loop is terminated when the progress is too small, that is when the objective does not progress more than $0.75\%$ every $6$ steps, or when $q$ reaches $q_{\max}=200$. We compute initial points $M_0$ using different representative tables as described in Algorithm~\eqref{algo:initial}. Figure~\ref{fig:initialpoints} illustrates the variation in performance for different choices of $M_0$. There is no ``natural'' distance between GIST features that we could consider. We have tried a few, taking for instance distances based on the $8$ directions and $4\times 4=16$ possible locations in the grid described by the $128$ GIST features, but we could not come up with one that was competitive with any of the methods considered above. This situation illustrates our claim in the introduction that GML can select agnostically a metric for features without using any prior knowledge on the features.

\begin{figure}
\BC\includegraphics[width=13cm]{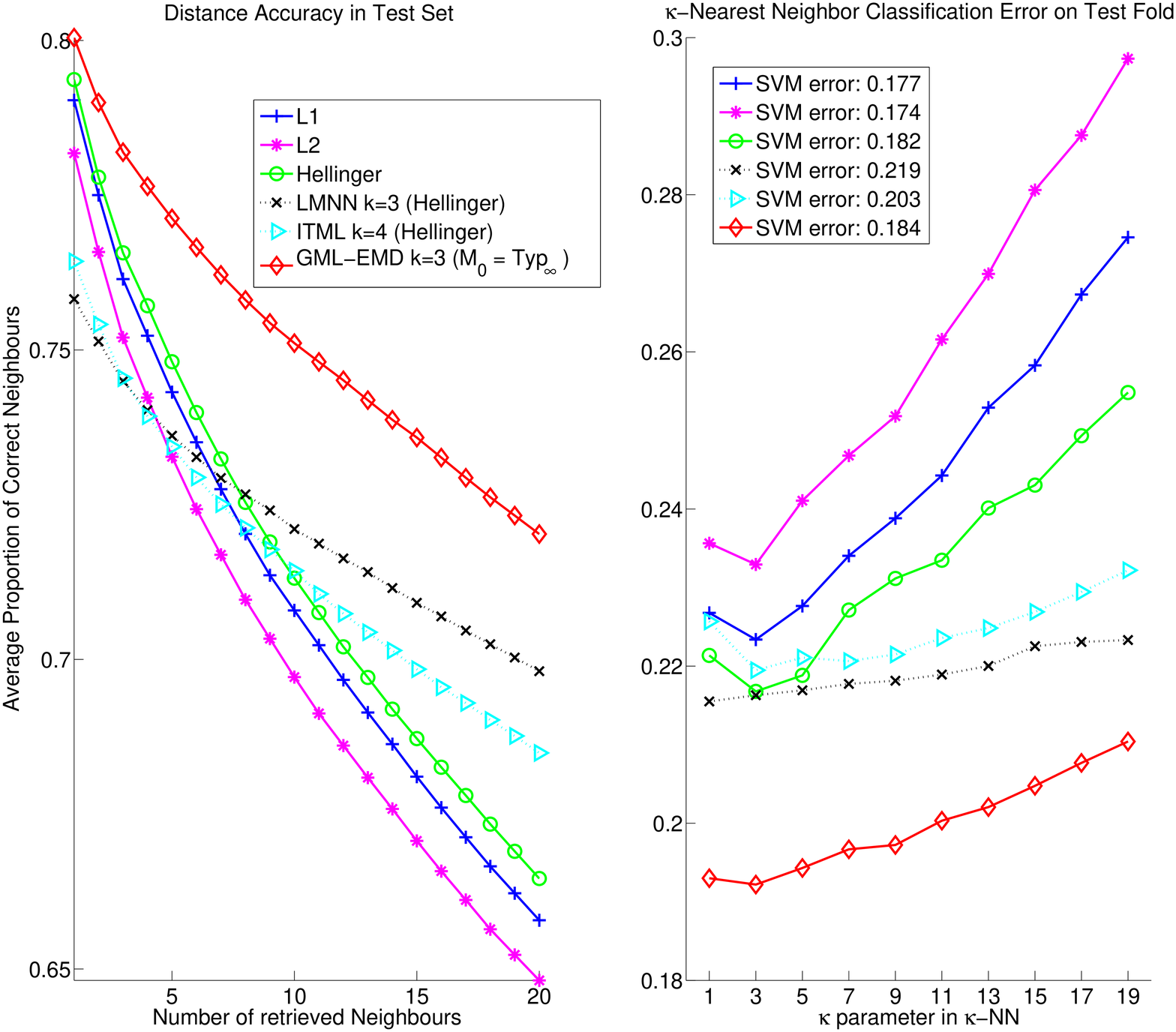}\EC
\caption{(left) Accuracy of each considered distance on the test set as measured by the average proportion, for each datapoint in the test set, of points coming from the same class within its $\kappa$ nearest neighbors. These proportions were averaged over $25\times 25=625$ classification problems using $40+40$ test points in each experiment. The ground metric in GML and Mahalanobis matrices in ITML and LMNN have been learned separately using a train set of $30+30$ points. (right) $\kappa$-NN classification error using the considered metrics and the $30+30$ points in the training fold, both to compute the metrics when needed and to compare test points. These results are also averaged for $40+40$ test points in each of the $625$ classification tasks. By $M_0=\text{Typ}_{\infty}$ we mean that the initial point was computed using typical tables and $k=\infty$ in Algorithm~\eqref{algo:initial}.\label{fig:results625}}
\end{figure}

\subsection{Comparison with the SVM baseline}
For each distance $d$, we consider its exponentiated kernel $\exp(-d/\sigma)$ and use a Support Vector Machine (SVM) classifier~\citep{Cortes1995} on the $80$ test points estimated with the $60$ training points. We use the following parameter grid to train the SVM's : the regularization parameter $C$ is selected within the range $\{1,10,100\}$; the bandwidth parameter $\sigma$ is selected as a multiple of the median distance $d$ computed in the training set, that is $\sigma$ is selected within the range $\{.1,.2,.5,1,2,5\}\times \text{med}\{d(r_i,r_j)\}$. We use a 4 folds (testing on left-out fold) and 2 repeats cross validation procedure on the training set to select the parameter pair that has lowest average error. Transportation distances are not negative definite in the general case, which is why we add a sufficient amount of diagonal regularization (minus the smallest negative eigenvalue) on the resulting $60\times 60$ Gram matrices to ensure that they are positive definite in the training phase.

\subsection{Results}
The most important results of this experimental section are summarized in Figure~\ref{fig:results625}, which displays , for all considered distances, their average recall accuracy on the test set and the average classification error using a $\kappa$-nearest neighbor classifier. These quantities are averaged over $N^2=25^2=625$ binary classifications. In this figure, GML used with EMD is shown to provide, on average, the best possible performance: the left hand figure considers test points and shows that, for each point considered on its own, GML-EMD selects on average more same class points as closest neighbors than any other distance. The performance gap between GML-EMD and competing distances increases significantly as the number of retrieved neighbors is itself increased. The right hand figure displays the average error over all $625$ tasks of a $\kappa$-nearest neighbor classification algorithm when considered with all distances for varying values of $\kappa$. In this case too, GML combined with EMD fares much better than competing distances. The average error when using a SVM with these distances is represented in the legend of the right-hand side figure. Our results agree with the general observation in metric learning that Support Vector Machines perform usually better than $\kappa$-nearest neighbor classifiers with learned metrics~\cite[Table 1]{weinberger2009distance}. Note however that the $\kappa$-nearest-neighbor classifier seeded with GML-EMD has an average performance that is directly comparable with that of support vector machines seeded with the $l_1$ or $l_2$ kernels. 

\begin{figure}
\BC\includegraphics[width=13cm]{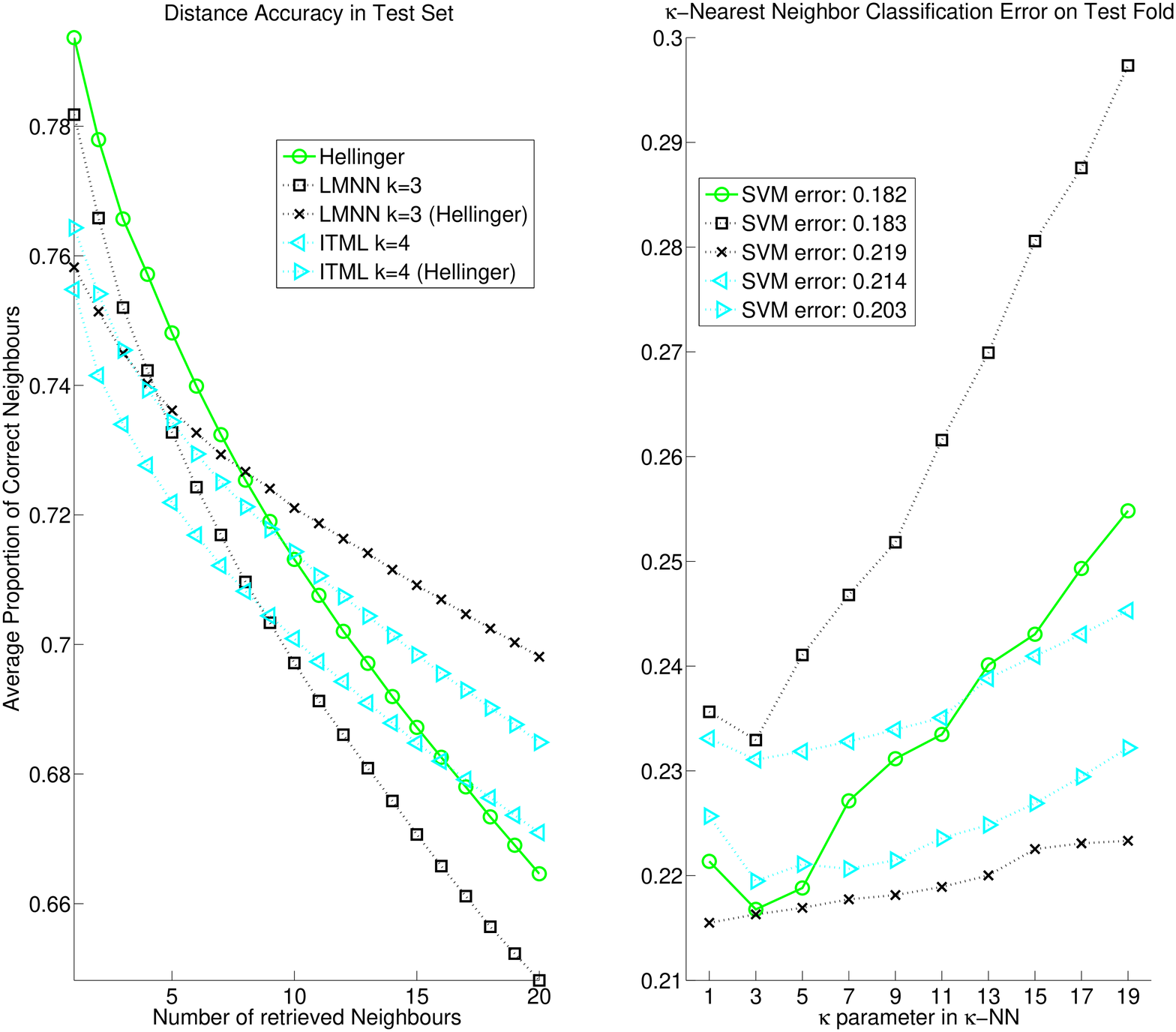}\EC
\caption{The experimental setting in this figure is identical to that of Figure~\ref{fig:results625}, except that only two different versions of LMNN and ITML are compared with the Hellinger distance. This figure supports our claim in Section~\ref{subsub:Mah} that Mahalanobis learning methods work better using the Hellinger representation of histograms, $\{\sqrt{r_i},\,i=1,\cdots,n\}$, rather than their straightforward representation in the simplex $\{r_i\}_{i=1,\cdots,n}$. \label{fig:lmnnanco}}
\end{figure}

It is also worth mentioning as a side remark that the $l_2$ distance does not perform as well as the $l_1$ or Hellinger distances on these datasets, which validates our earlier statement that the Euclidean geometry is usually a poor choice to compare histograms directly. This intuition is further validated in Figure~\ref{fig:lmnnanco}, where Mahalanobis learning algorithms are show to perform significantly better when they use the Hellinger representation of histograms. 

\begin{figure}
\BC\includegraphics[width=13cm]{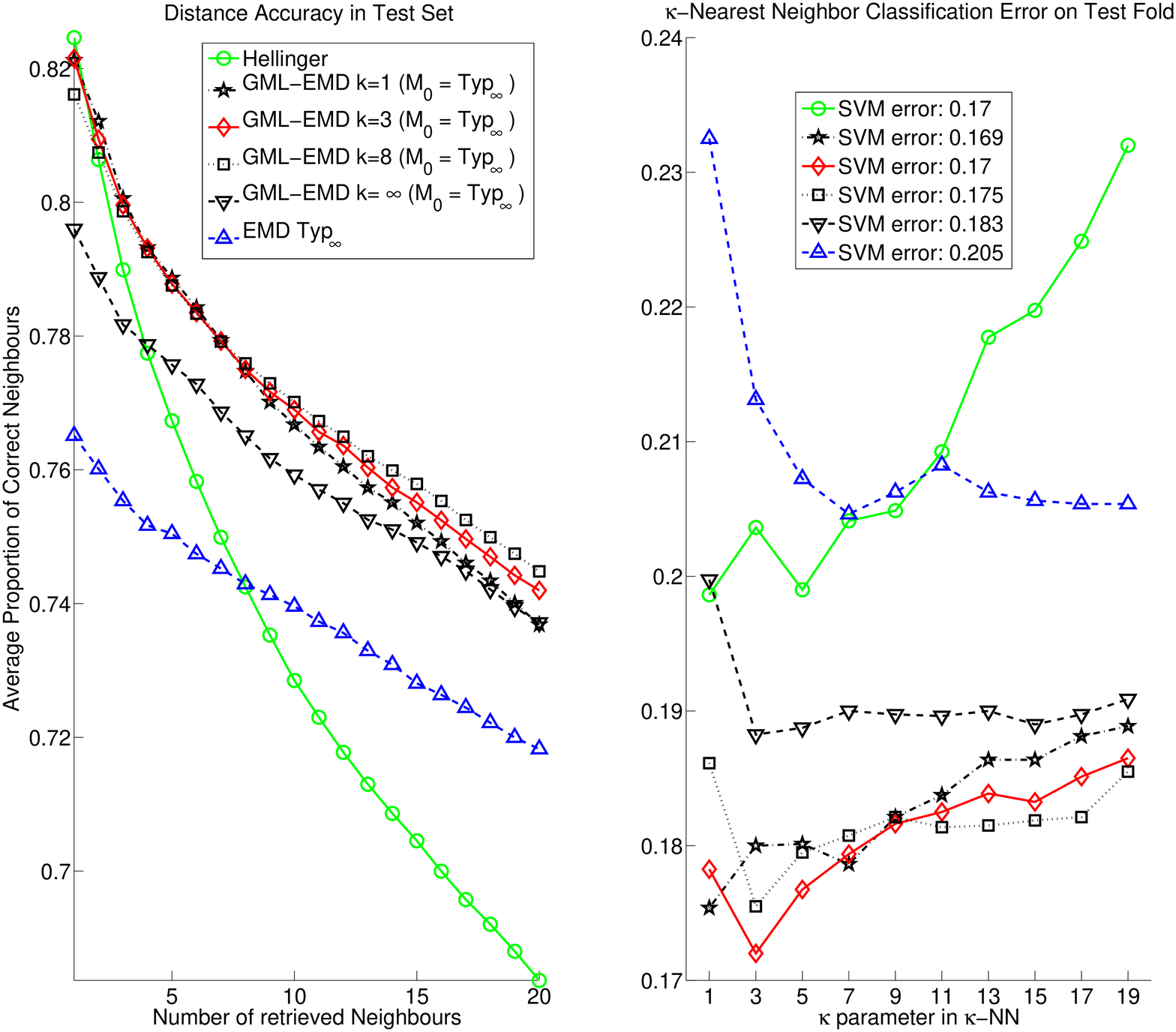}\EC
\caption{Distance accuracy and $\kappa$-nearest neighbor classifier error for GML-EMD using different values for the neighborhood parameter $k$. The initialization setting $M_0=\text{ Typ}_{\infty}$ is explained in the caption of Figure~\ref{fig:results625}. The last curve, \texttt{EMD Typ$_{\infty}$} displays the results corresponding to the EMD with a ground metric set directly to the output of Algorithm~\eqref{algo:initial} using typical tables and $k=\infty$. The difference in performance between these curves and that of \texttt{GML-EMD $k=3, (M_0=\text{ Typ}_{\infty})$} illustrates the progress achieved by Algorithm~\ref{algo:optim}. The performance curves also agree with the intuition that the GML metric set at a given neighborhood parameter $k$ has a comparative advantage over other GML metrics when the $\kappa$ parameter of nearest neighbor classifiers is itself close to $k$. The experimental setting is identical to that of Figure~\ref{fig:results625}, except that experiments were averaged here over $10\times 10=100$ classification problems, instead of $625$. There is no overlap between these two sets of binary classifications.\label{fig:manyks}}
\end{figure}

Figure~\ref{fig:manyks} shows that the performance of GML can vary significantly depending on the neighborhood parameter $k$. We have also considered as a ground metric the initial point $M_0=\text{Typ}_{\infty}$ obtained by using typical tables and $k=\infty$ with Algorithm~\ref{algo:initial}. The corresponding results appear as \texttt{EMD Typ$_{\infty}$} curves in the figure. 
These curves are far below those corresponding to \texttt{GML-EMD $k=3 \;(M_0=$Typ$_{\infty})$}. This gap illustrates the fact that the subgradient descent performed by Algorithm~\ref{algo:optim} does have a real impact on performance, and that choosing a good initial point in itself is not enough. 

\begin{figure}
\BC\includegraphics[width=13cm]{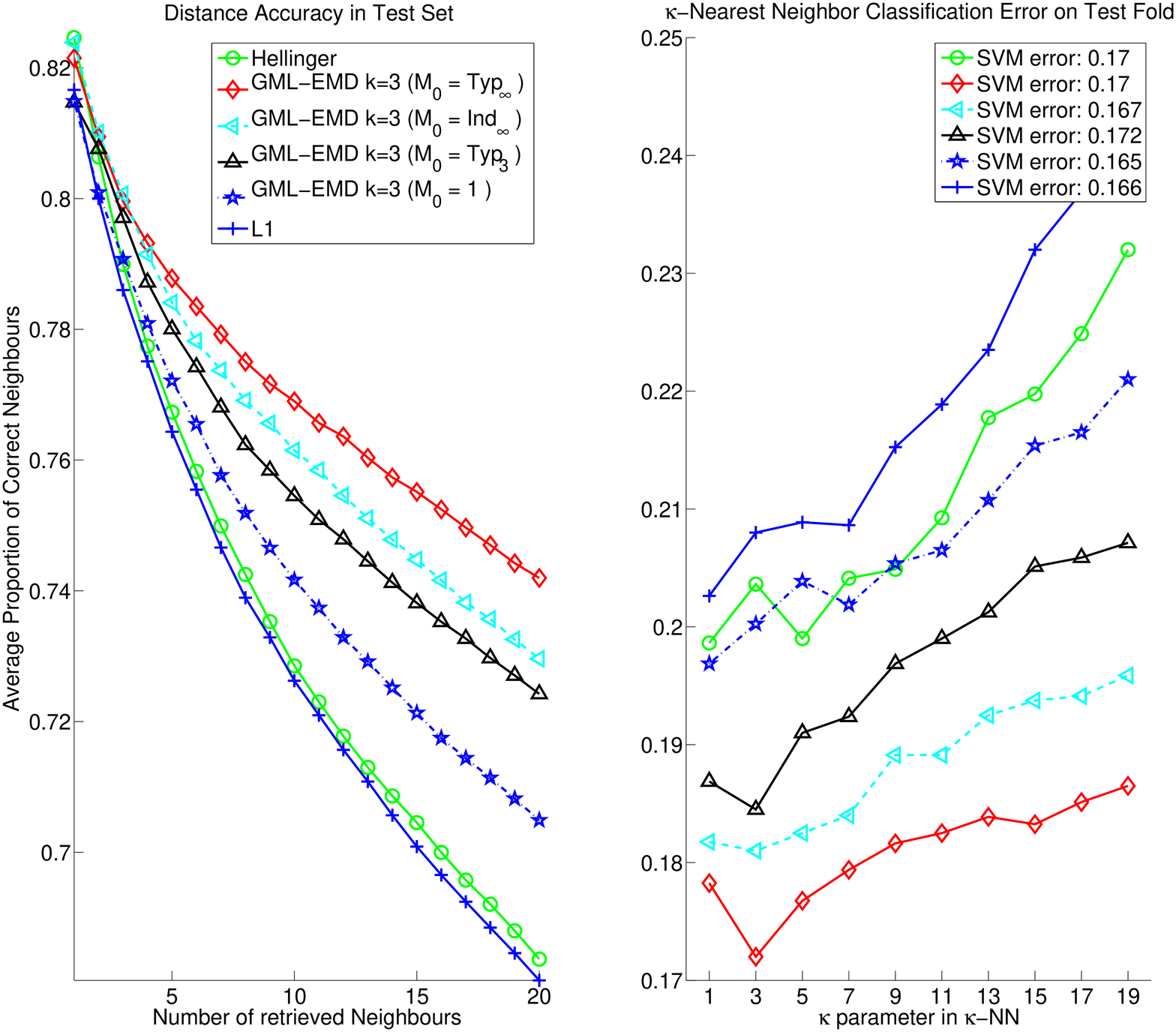}\EC
\caption{Distance accuracy and $\kappa$-nearest neighbor performance for GML-EMD set at $k=3$ using different initial points described in Section~\ref{sec:initial}, and particularly Section~\ref{subsub:barv}. This figure shows that, despite its computational overhead, initializing Algorithm~\ref{algo:optim} using typical tables performs better on average than using independence tables. The experimental setting is identical to that of Figure~\ref{fig:manyks}.\label{fig:initialpoints}}
\end{figure}

Finally, Figure~\ref{fig:initialpoints} reports additional performance curves for different initial points $M_0$. These experiments tend to show that, despite their computational overhead,~\citeauthor{barvinok2009asymptotic}'s typical tables seem to provide a better initial point than independence tables in terms of average performance. Please note that $N=25$ in Figures~\ref{fig:results625} and~\ref{fig:lmnnanco}, and $N=10$ in Figures~\ref{fig:manyks} and~\ref{fig:initialpoints}.

\section{Conclusion and Future Work}\label{sec:conclusion}

\subsection{Overview}
We have proposed in this paper an approach to tune adaptively the unique parameter of transportation distances, the ground metric, given a training dataset of histograms. This approach can be applied on any type of features, as long as a set of histograms along with side-information, typically labels, are provided for the algorithm to learn a good candidate for the ground metric. The algorithms performs a projected subgradient descent on a difference of convex functions, and can only find local minimizers. We propose a few initial points to compensate for this arbitrariness, and show that our approaches provide, when compared to other competing distances, a superior average performance for a large set of image binary classification tasks using GIST features histograms.

\subsection{Ground Metric and Computational Speed}
We have argued in the introduction of this paper that the ground metric was never considered so far as a parameter that could be learned from data. The ground metric has however attracted a lot of attention recently for a different reason.~\citet{ling2007efficient,gudmundsson2007small,Pele-iccv2009,ba2011sublinear} have all recently argued that the computation of the EMD can be dramatically sped up when the ground metric matrix has a certain structure. For instance, ~\citet{Pele-iccv2009} have shown that the computation of each earth mover's distance can significantly reduced whenever the larger values of any arbitrary ground metric are thresholded to a certain level. Ground metrics that follow such constraints are attractive because they result in transportation problems which are provably faster to compute. Our work in this paper suggests on the other hand that the content (and not the structure) of the ground metric can be learned to improve classification accuracy. We believe that the combination of these two viewpoints could result in transportation distances that are both adapted to the task at hand and fast to compute. A strategy to achieve both goals would be to enforce such structural constraints on candidate metrics $M$ when looking for minimizers of criteria $C_k$.

The mot computationally intensive part of Algorithm~\eqref{algo:optim} lies in the repeated calls to Algorithm~\eqref{algo:sub} which itself computes a number $O(n^2)$ optimal transportations between pairs of histograms, where $n$ is the size of the training set. We have used the network simplex algorithm with warm starts to carry out these computations, but faster alternatives along the lines of the implementations provided by~\citet{Pele-iccv2009} could provide computational improvements. The EMD is also known to accept lower bounds in particular cases. An efficient lower bound would reduce considerably the complexity of Algorithm~\eqref{algo:sub} by narrowing down the computation of Optimal transportations to a smaller subset of neighbor candidates.

\subsection{Future Applications}
We have proposed in this paper a set of techniques to learn ground metrics using histograms of arbitrary features. We believe these techniques will prove useful to study histograms of latent features, such as topics ~\cite{blei2009topic} or Dictionaries~\citep{kreutz2003dictionary,mairal2009online,jenatton2010proximal}, for which natural metrics are not always available.

\bibliography{bib_short}
\bibliographystyle{plainnat}
\end{document}